\documentclass{article}
\usepackage{spconf,amsmath,graphicx}
\usepackage{etoolbox}
\usepackage{cite}
\usepackage{subfigure}
\usepackage{comment}
\usepackage{booktabs}
\usepackage{multirow}
\usepackage{setspace}
\usepackage{color}
\usepackage[linesnumbered,ruled,vlined]{algorithm2e}
  
\title{Spread Control Method on Unknown Networks Based on Hierarchical Reinforcement Learning}
\name{Wenxiang~Dong, Zhanjiang~Chen, H.~Vicky~Zhao}
\address{Department of Automation, Tsinghua University, Beijing, 100084, China}

\begin{document}
%
\maketitle

\begin{abstract}
Epidemics such as COVID-19 pose serious threats to public health and our society, and it is critical to investigate effective methods to control the spread of epidemics over networks. Prior works on epidemic control often assume complete knowledge of network structures, a presumption seldom valid in real-world situations. In this paper, we study epidemic control on networks with unknown structures, and propose a hierarchical reinforcement learning framework for joint network structure exploration and epidemic control. To reduce the action space and achieve computation tractability, our proposed framework contains three modules: the Policy Selection Module, which determines whether to explore the structure or remove nodes to control the epidemic; the Explore Module, responsible for selecting nodes to explore; and the Remove Module, which decides which nodes to remove to stop the epidemic spread. Simulation results show that our proposed method outperforms baseline methods.
\end{abstract}
\section{Introduction}


The recent COVID-19 pandemic has resulted in a substantial number of infections and extensive societal and economic losses. Additionally, the rampant spread of misinformation, rumors, and malicious content on the internet has exacerbated the situation. All of the above can be modeled as the propagation of disease/information in complex and dynamically changing networks. It is crucial to investigate how to manage the spread of detrimental events within these complex networks. In this work, we use epidemic control as an example and study effective mechanisms to hinder the dissemination of such harmful events. Our work can also be extended to other scenarios, such as stopping the propagation of rumors.


Prior works in \cite{2015Node,tong2010vulnerability,xiong2017effective,prakash2012threshold,2010Virus,2015Node2} propose to control the spread of epidemics by removing nodes from the network. One simple approach for selecting nodes to remove is based on their degrees in the network \cite{2015Node}. Additionally, as the network’s ability to resist the spread of the disease is related to the largest eigenvalue of its adjacency matrix \cite{2003Epidemic}, many works tried to remove nodes to alter the network's largest eigenvalue. In \cite{tong2010vulnerability,2015Node2}, the ``Shield-value" was used to quantify the reduction in eigenvalues resulting from node removal. Furthermore, the works in \cite{2017Spectral, 2018Group} proposed ``GreedyDrop'' and ``SVID'', respectively, to decrease the maximum eigenvalue of the adjacency matrix and to control the epidemic.
Moreover, some works explored the strategies for curbing the spread of diseases by managing population movements on a larger scale, such as within different regions \cite{wan2021multi,hota2021closed,fajgelbaum2021optimal}.


The above works assume complete knowledge of the network structure; while in reality, the network structure is often unknown due to the complexity, dynamic nature, and difficulties in data acquisition.
Many prior works studied how to find missing links in networks with unknown structures. In \cite{lu2011link,kumar2020link}, the authors summarized and introduced existing link prediction methods, including probabilistic methods, similarity-based methods, dimensionality reduction approaches, and others. However, they only study how to find missing links and do not consider how to control the propagation of the epidemic.

During public health crises such as COVID-19, government authorities frequently face the challenge of managing the disease using measures such as quarantining infected individuals and those at high risk. They must also discover unknown links and identify close contacts of infected individuals, all while working within the constraints of limited resources, such as manpower. In this work, we study the challenging task of controlling the epidemic over networks with unknown structures and investigate joint epidemic control and network structure exploration. We consider the scenario with limited resources, and at each time, we assume that we can choose either to explore neighbors of $M_1$ nodes or to remove $M_2$ nodes to stop the epidemic. We propose a HIerarchical Reinforcement learning framework for Epidemic Control ({HIREC}) to learn the optimal policy. To reduce the computation complexity, our proposed framework first chooses between the two actions of explore and remove, and then decides which nodes to explore/remove. Simulation results validate its effectiveness. 

\begin{figure*}[htpb]
	
	\centering
	\includegraphics[width=0.85\linewidth]{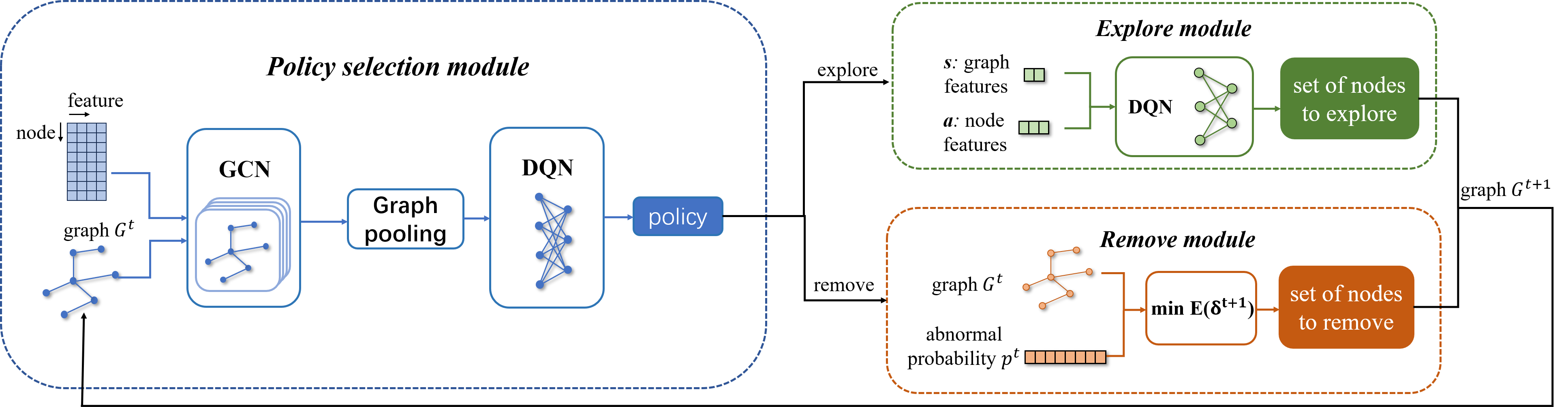}
	\centering \vspace{-3mm}
	\caption{The Proposed Hierarchical Reinforcement Learning Framework for Epidemic Control ({HIREC}).
	}
	\label{fig:model}
\end{figure*}
\section{Problem Formulation}

In the epidemic control example, we consider a network with a total of $N$ nodes, where each node represents a user and an edge connecting two nodes indicates that there is physical contact between the two corresponding users.
We assume that the nodes are known and the edges are unknown initially at time $t=0$. We assume that an epidemic is spreading over the network, and we use the classic SEIR model \cite{li1995global} to model the spread of the disease.
Each node $i$ in the system can be in one of four health states: \emph{Susceptible} (S), \emph{Exposed} (E), \emph{Infected} (I), or \emph{Recovered} (R). \emph{Susceptible} nodes are individuals who have not yet been infected and are susceptible to the disease.
\emph{Exposed} nodes are individuals who have been exposed to the disease but have not yet shown symptoms.
\emph{Infected} nodes are individuals who have shown symptoms. 
\emph{Recovered} nodes are individuals who have recovered from the disease and have gained permanent immunity. Both exposed and infected individuals may spread the disease to susceptible users, and we say a node is ``abnormal'' when the user is in the $E$ or $I$ state. When a susceptible node comes into contact with an abnormal node, it has a probability $\beta$ of becoming exposed. The exposed node then progresses to the infected state and shows syndromes after a fixed time period $t_0$. Additionally, at each time step, an infected node recovers with probability $\gamma$ and gains permanent immunity.

In real-world scenarios, policymakers often face the challenge of controlling the spread while simultaneously exploring the underlying network structure. To stop the further spread of the disease, individuals with high probabilities of being abnormal need to be quarantined, where all links, both known and unknown, connected to them are removed. When they recover or the epidemic ends, all their previous links are reconnected. Meanwhile, we need to explore the network structure to know the current connections between individuals. This can help us find close contacts of those with a high risk of being abnormal \cite{christley2005infection}.
We define these two actions as \emph{remove} and \emph{explore}, respectively.
We consider the scenario where there are limited resources available and assume that at each time, the policymaker can choose to either explore neighbors of $M_1$ nodes or remove $M_2$ nodes with high risk, but not both. Due to exploration and removal, the network structure we can observe will change at each time step, and we use $\mathcal{G}^t$ to represent the observable network at time $t$, and $G^t$ is its adjacency matrix. 


Our goal is to minimize the abnormal rate $r_a(T)$, the percentage of nodes who have been in state $E$ or $I$ before time $T$. 
We use two binary $N\times 1$ vectors to represent the policy at time $t$, where $a_i^t=1$ indicates that node $i$ is chosen for exploration at time $t$, and $b_i^t=1$ indicates that node $i$ is chosen for removal at time $t$. 
Note that due to the resource constraint mentioned earlier, at time $t$, either $M_1$ entries in $\pmb{a}^t$ are set to 1 and $\pmb{b}^t$ is an all-zero vector, or $M_2$ entries in $\pmb{b}^t$ are set to 1 and $\pmb{a}^t$ is all zero.
Then, the problem becomes \vspace{-2mm}
\begin{align}
\label{eq:o-problem}
&\underset{\pmb{a}^1,\pmb{a}^2,...,\pmb{a}^{T},\pmb{b}^1,\pmb{b}^2,...,\pmb{b}^{T}} {\mathbf{min}} r_a(T),\\
s.t.&\ (||\pmb{a}^t||_0=M_1,||\pmb{b}^t||_0=0) \text{ or } (||\pmb{a}^t||_0=0, ||\pmb{b}^t||_0=M_2),\nonumber \\
&\pmb{a}^t\in \{0,1\}^N ,\pmb{b}^t\in \{0,1\}^N ,t=1,2,...,T,  \nonumber \vspace{-2mm}
\end{align}
where $||\pmb{x}||_0$ is the $l_0$ norm of $\pmb{x}$.

The challenges in solving the above problem are: first, due to the existence of the unobservable exposed state, only part of the abnormal nodes can be observed. To address this issue, we use the abnormal probability estimation module proposed in our prior work to estimate each node's probability of being in state $E$ or state $I$, the details are in Section 3.1. Secondly, since the network structure is only partially observable, we cannot obtain an accurate estimate of $r_a(T)$.
A possible solution is to use reinforcement learning methods to find the optimal policy. However, at each time step, there are $\binom{N}{M_1}+ \binom{N}{M_2}$ possible actions, which makes it extremely time-consuming
for traditional reinforcement learning methods. To address this challenge, we employ a divide-and-conquer approach and propose a hierarchical reinforcement learning framework to reduce the action space.


\section{The proposed HIREC framework}
In our proposed HIREC framework shown in Fig.\ref{fig:model}, at time $t$, given the current known network structure and the estimated abnormal probabilities, the Policy Selection Module decides whether to adopt the explore or the remove strategy. Then depending on the selected policy (explore or remove), either the Explore Module selects which $M_1$ nodes to explore, or the Remove Module determines which $M_2$ nodes to remove. 
In the following, we will introduce the details of all modules.



\subsection{Abnormal Probability Estimate Module}
The Abnormal Probability Estimate Module estimates the abnormal probability of each node in the network based on the known node states and network structure. This estimated probability is then used as a feature input into the other three modules.

Based on the observable network $\mathcal{G}^t$ at time $t$ and the individuals in state $I$, we try to estimate the probability of each node on the network being abnormal. Define $p_i^t$ as the abnormal probability that the node $i$ is in state $E$ or $I$. Since state $I$ is observable, we use $q_i^t \in \{0,1\}$ indicate whether node $i$ has been in state $I$ before. For the convenience of representation, we use $H_i^t$ to represent the state of node $i$ at time $t$. For example, $H_i^t=S$ means that node $i$ is in state $S$ at time $t$.

From the SEIR model, node $i$ is in state E or state I at $t+1$ if:
\begin{itemize}
	\item $i$ is state $E$ at time $t-1$,
	\item $i$ is state $I$ at time $t-1$, and has not recovered at $t$,
	\item $i$ is at state $S$ at time $t-1$, is infected by his/her neighbors and becomes exposed at $t$.
\end{itemize}

Then we have:
\begin{align}
p_{i}^{t}&=P \left[H_i^{t-1}=E \right] + P \left[H_i^{t}=I | H_i^{t-1} = I \right] P[H_i^{t-1} = I] \nonumber \\& \quad + P \left[H_i^{t}=E | H_i^{t-1} = S \right] P[H_i^{t-1} = S]\label{con:0}\\
&=P \left[H_i^{t-1}=E \right] + P \left[H_i^{t}=I | H_i^{t-1} = I \right] P[H_i^{t-1} = I] \nonumber \\& \quad + \left( 1-P[H_i^{t} =S |H_i^{t-1} =S] \right) P[H_i^{t-1} = S]\nonumber\\
&= p_{i}^{t-1} q_{i}^{t-1} + p_{i}^{t-1} (1-q_{i}^{t-1})(1-\gamma) \nonumber\\ & \quad +
[1-p_{i}^{t-1}] q_{i}^{t-1} [1-\prod_{j=1...N}(1-G^{t-1}_{i,j}\beta p_{j}^{t-1})] \nonumber\\&
=\left(1-\gamma (1-q_{i}^{t-1})\right)p_{i}^{t-1}\nonumber\\
&\quad +[1-p_{i}^{t-1}] q_{i}^{t-1} [1-\prod_{j=1...N}(1-G^{t-1}_{i,j}\beta p_{j}^{t-1})],\nonumber
\end{align}

If we know the $p_i^{t-1}$, then we can get the $p_i^t$ by $p_i^{t-1}$

\begin{align}
p_{i}^{t}=&[1-p_{i}^{t-1}] q_{i}^{t-1} [1-\prod_{j=1...N}(1-\beta G_{i,j}^{t-1} p_{j}^{t-1})]\nonumber\\
&+\left(1-\gamma (1-q_{i}^{t-1})\right)p_{i}^{t-1},\label{eq:p}
\end{align}
where $G_{i,j}^{t-1}=1$ when $i$ and $j$ have connection in observable network $\mathcal{G}^{t-1}$, and otherwise $G_{i,j}^{t-1}=0$.

Then how could we get the $p_i^{t-1}$? On the one hand, $p_i^{t-1}$ can be obtained recursively from $p_i^{t-2}$. On the other hand, if the node $i$ enters state $I$ at time $t$, since the incubation period is $t_0$, then it has been abnormal since $t-t_0$ and we have $p_i^{t-t_0}=p_i^{t-t_0+1}=...=p_i^{t}=1$. In addition, when new nodes are explored, the previously obtained $p_i^{1}...p_i^{t-1}$ may be inaccurate because the network structure is updated. Therefore, at each moment, we revise the past estimates $p_i^{1}...p_i^{t-1}$ according to the current new information (the new infected nodes and the new observable network), and recursively obtain the current probability $p_i^{t}$. The complete model is in Algorithm \ref{algor:estimate}.

\begin{algorithm}
	\SetAlgoLined
	\caption{Abnormal Probability Estimate}
	\label{algor:estimate}
	$p^1_0=p^1_1=...=p^1_N=0$
	
	\For{$t=2\text{ to }T$}
	{
		$\tilde{t} = \max \{ 1,t-t_0 \}$

		\For{$\tau=\tilde{t}\text{ to }t-1$}
		{
			
		$p_i^{\tau}=1$, for all node $i$ in the new infected set.
		
		Update $\mathcal{G}^{\tau}$ based on the new explore nodes.
		
		Get $p_i^{\tau+1}$ by $p_i^{\tau}$ based on \eqref{eq:p}.

	}
		
		Get $p_i^t$ for all node $i$ and output.
		
	}
	
\end{algorithm}

\subsection{The Policy Selection Module}
The policy selection module is responsible for making decisions on whether to explore or remove nodes based on the current state of the network and the current known health states of all nodes. For this purpose, we utilize the Deep Q Network (DQN) framework \cite{mnih2015human} to learn the optimal policy. To apply DQN effectively, the policy selection task needs to be formally defined as a Markov Decision Process (MDP) with states, actions, and rewards.

\textbf{State}:  
The state includes the entire network's current condition, which is composed of two components: the observable network $\mathcal{G}^t$ and the node feature matrix $F^{0,t} \in R^{N\times 4}$, where $N$ represents the number of nodes in the network. 
The features of node $i$ include $p_i^t$ (the estimated probability of node $i$ being abnormal at time $t$), its explore state (equals 1 if the node has been explored, and 0 otherwise), its remove state (equals 1 if the node is currently removed, and 0 otherwise), and its degree in the observable network. 

However, if $\mathcal{G}^t$ and $F^{0,t}$ are directly inputted into DQN, the model is related to the graph's size $N$ and needs to be retrained whenever it changes. To achieve effective learning and enable training and deployment on networks of different sizes, we use the Graph Convolutional Network (GCN) \cite{kipf2017semisupervised} and a max pooling layer to obtain the state embeddings. GCN refines node features within a graph by propagating and aggregating information from neighboring nodes. Given an $M$-level GCN, from \cite{kipf2017semisupervised}, for the $k$th layer, we have
\begin{equation}
	F^{k,t} = ReLU(\tilde{D}^{-\frac{1}{2}}\tilde{G^t}\tilde{D}^{-\frac{1}{2}}F^{k-1,t}W_{k-1}),
\end{equation}
where $F^{k,t}$ and $F^{k-1,t}$ are the matrices of node features in the $k$th and $(k-1)$th layers, respectively, and $W_{k-1}$ is the weight matrix of the $(k-1)$th layer. $\tilde{G^t} = G^t+I$, where $G^t$ is the adjacency matrix of $\mathcal{G}^t$ and $I$ is the identity matrix. $\tilde{D}$ is a diagonal matrix with $\tilde{D}_{i,i} = \sum_j \tilde{G}^t_{i,j}$. Then, given the $M$th layer features $F^{M,t}$, we use global pooling to obtain the graph representation $s^t$, which serves as the state embedding and is inputted into DQN to make informed decisions. 


\textbf{Action}: There are two possible actions at each time step in this task, explore and remove.

\textbf{Reward}: 
We set the reward $r^t=-\delta^t$, where $\delta^t$ is the number of susceptible nodes that are exposed to the disease and transit from state $S$ to state $E$ at time $t$. However, since exposed nodes show no symptoms and are not observable. 
Note that we assume that the incubation period $t_0$ is fixed, and those $\delta^t$ newly exposed individuals will show syndromes and become newly infected at time $t+t_0$. We let $\delta^t$ equal to the number of newly infected nodes at time $t+t_0$. 
\vspace{-3mm}
\subsection{The Explore Module}
The explore module decides which nodes to explore, and we use DQN to learn the optimal strategy. To reduce the action space and the computation complexity, instead of selecting $M_1$ nodes all at once, it selects nodes to explore one by one, and this process is repeated $M_1$ times. The states, actions, and rewards of the explore task are as follows.


 
\textbf{State}: We represent the state using a 2-element vector. The first element is the proportion of nodes already explored by time $t$, reflecting the progress made in the network structure exploration task. The second one is the proportion of infected nodes at time $t$, indicating the severity of the epidemic.

\textbf{Action}: The action space is the node set. To represent each node as an action input in the DQN, we use the node's features. Node $i$'s action representation is a vector of length 3 that includes $p_i^t$ 
(the estimated probability that node $i$ is abnormal at time $t$), its explore state (equals 1 when it has been explored, and 0 otherwise), and its degree in the observable network. 
This representation is then used as input to the DQN to make informed decisions on node selection for exploration.

\textbf{Reward}: 
Intuitively, the exploration task should discover as many new edges as possible, especially those connected to the known infected nodes. This is because neighbors of infected nodes often have a higher probability of being abnormal, and finding these nodes can help us better identify nodes to remove later. If we choose node $k$ to explore at time $t$, then we define the reward as $r_k^t=\rho\varphi_k^{t}+\xi_k^{t}$, where $\varphi_k^{t}$ is the number of newly discovered edges connected to currently infected nodes after exploring node $k$ at time $t$, and $\xi_k^{t}$ is the number of newly discovered edges connected to nodes that are not in state \emph{I} after exploring node $k$ at time $t$. $\rho$ is a hyperparameter to quantify the importance of finding close contacts of infected nodes. 

\subsection{The Remove Module}

Given the current known information of the network and the health states, the remove module selects $M_2$ nodes to remove to effectively control the epidemic spread.  
Similar to the explore module, to reduce the computation complexity, we select nodes to remove one by one and repeat this process $M_2$ times. 
Since state $I$ is observable, we define $q_i^t \in \{0,1\}$, where $q_i^t=0$ means node $i$ has been in state $I$ before time $t$ and $q_i^t=1$ otherwise. In each iteration, we find the optimal node to remove to minimize $E(\delta^{t+1})$, the expected number of newly exposed nodes who transit from state $S$ to state $E$ at time $t+1$. Then we have Theorem 1.

\begin{figure*}[tbp]
	\centering
	\subfigure[$M_2=1$]{
		\begin{minipage}[t]{0.33\linewidth}
			\centering
			\includegraphics[width=0.9\linewidth]{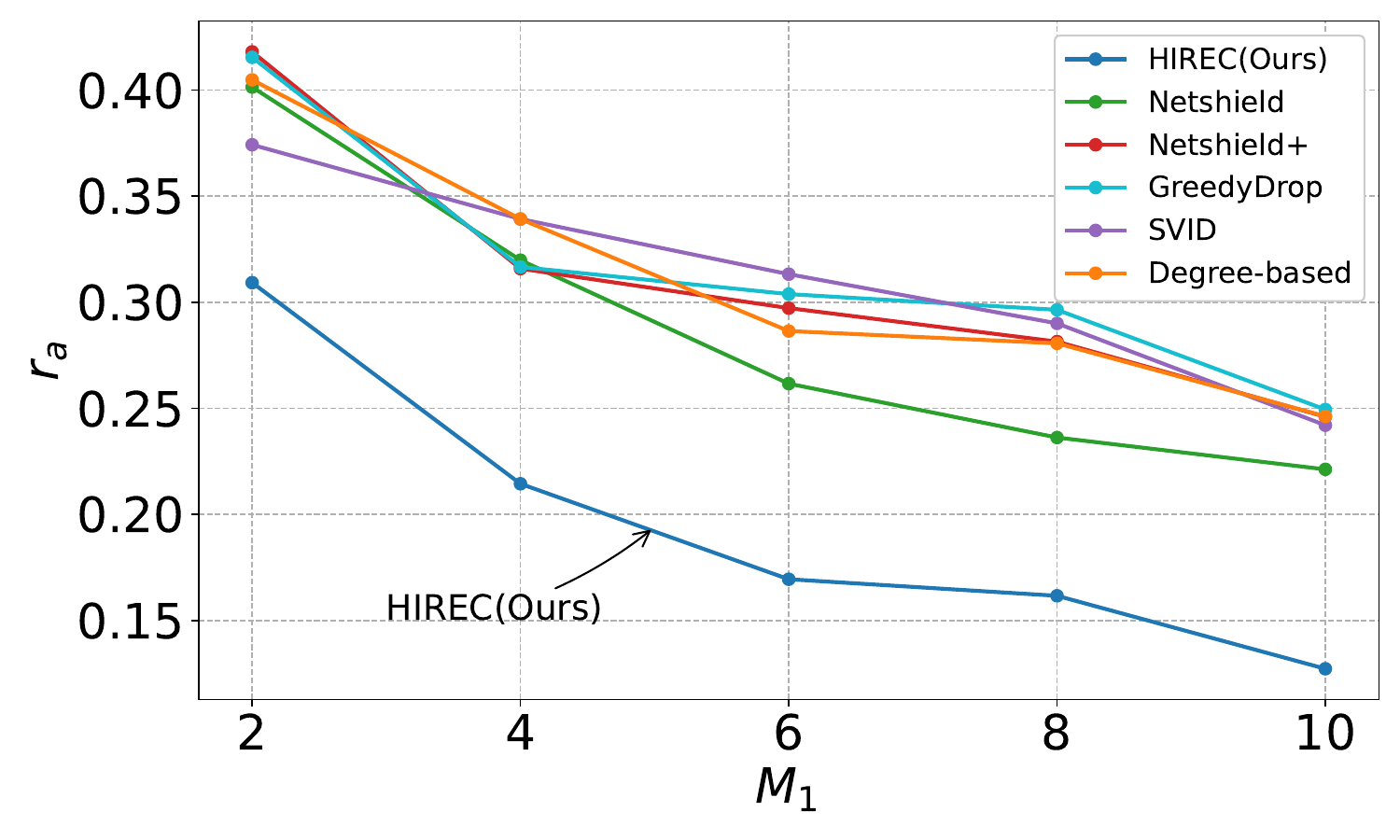}
			\label{fig:M1_1}
		\end{minipage}%
	}%
	\subfigure[$M_2=2$]{
		\begin{minipage}[t]{0.33\linewidth}
			\centering
			\includegraphics[width=0.9\linewidth]{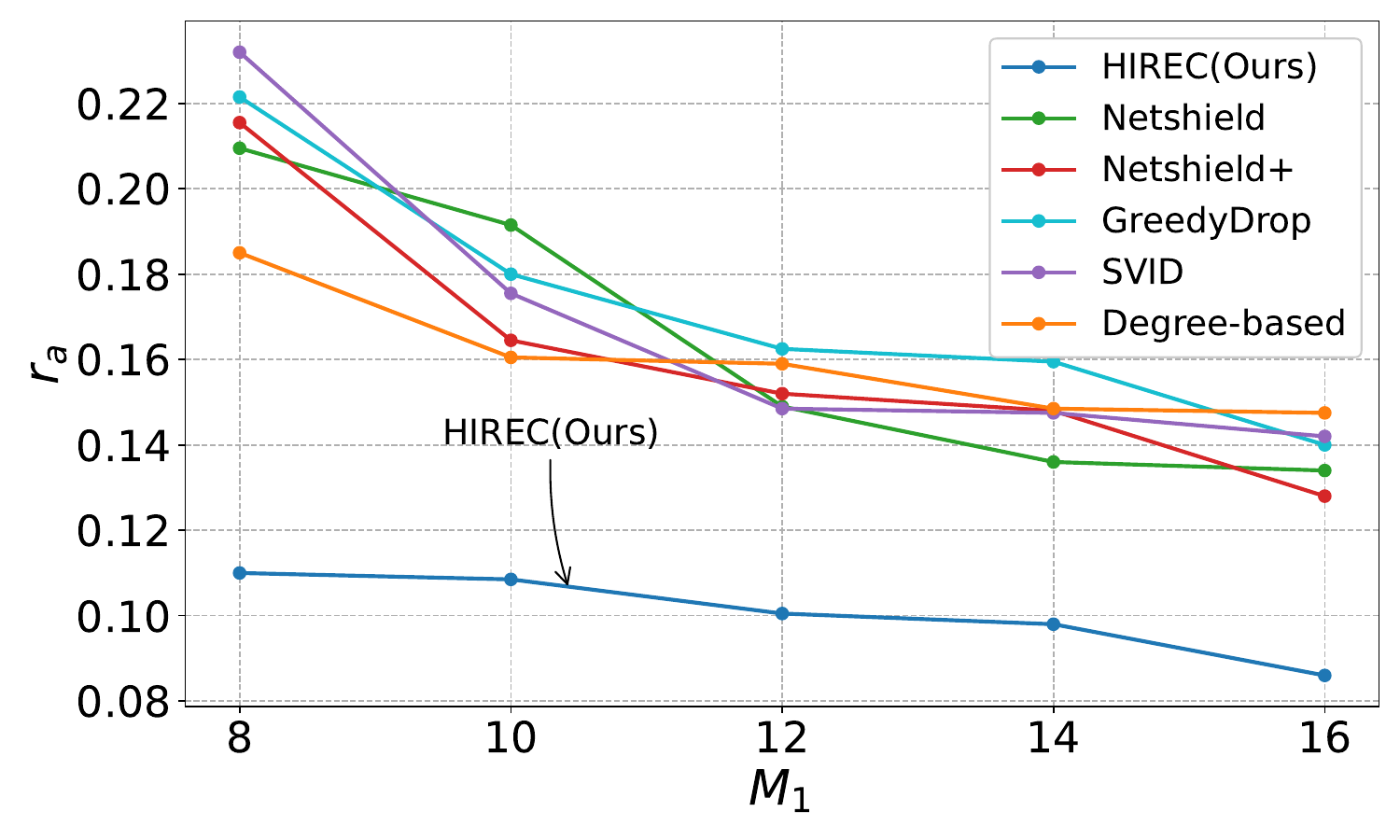}
			\label{fig:M1_2}
		\end{minipage}%
	}%
	\subfigure[]{
		\begin{minipage}[t]{0.36\linewidth}
			\centering
			\includegraphics[width=0.9\linewidth]{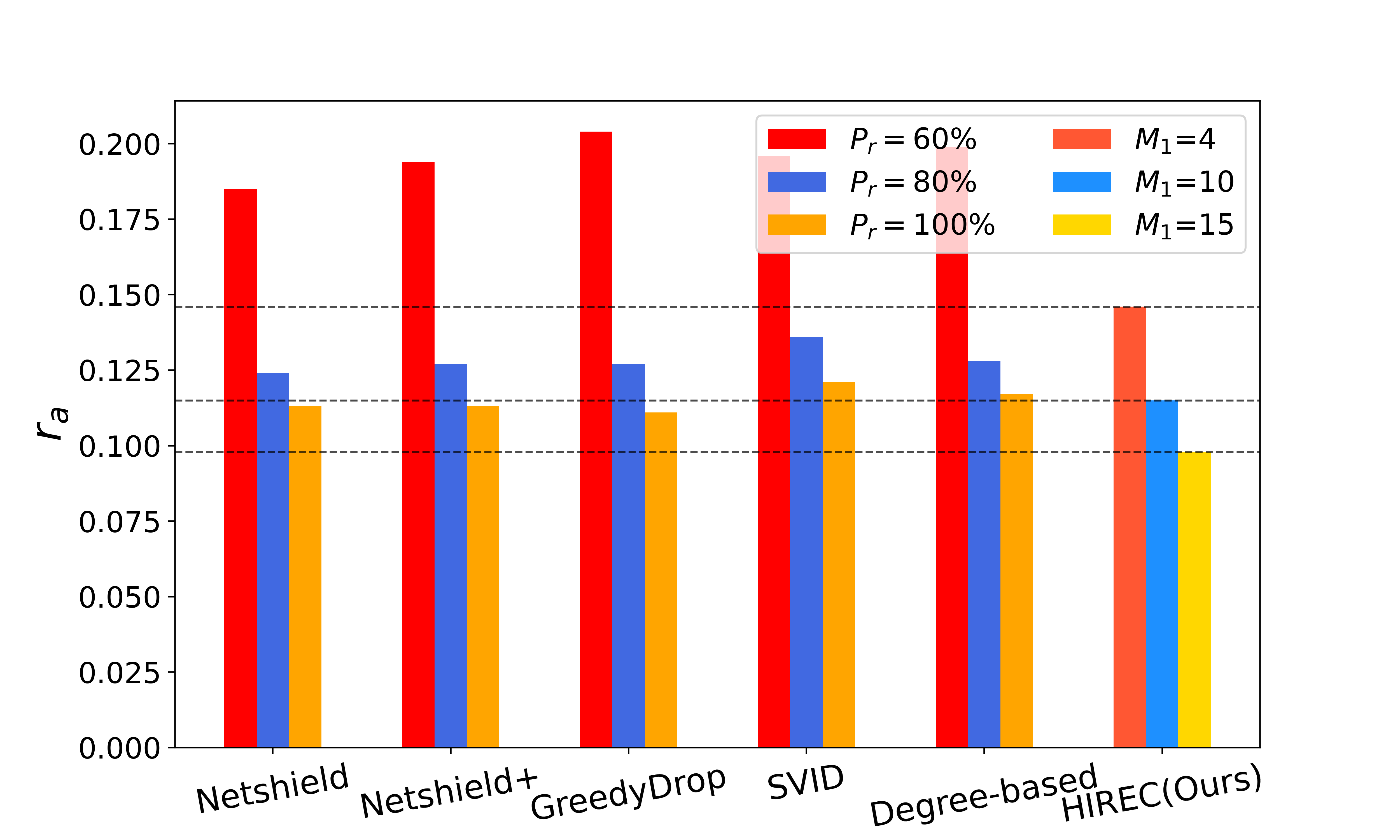}
			\label{fig:s2}
		\end{minipage}%
	}%
	\centering \vspace{-3mm}
	\caption{Performance comparison between HIREC and the baseline methods. (a) and (b): The network structure is unknown to all methods. (c) The baseline methods know a fixed part of the network structure while HIREC has no such information.
	}
	\label{Algor_Compare}
\end{figure*}

\noindent\textbf{Theorem 1} \emph{Finding the optimal node to remove at $t$ to minimize $E(\delta^{t+1})$ is equivalent to finding node $k$ with the highest $f(k)$, where} \vspace{-2mm}
	\begin{align}
	&f(k)=(1-p_{k}^t) q_{k}^t [1-\prod_{i=1...N}(1-G^t_{k,i}\beta p_{i}^t)] \label{eq:removeori}\\
	&+\sum_{i=1...N,i\neq k}G^t_{i,k}\beta p_{k}^t(1-p_{i}^t)q_{i}^t\prod_{j=1...N,j\neq k}(1-G^t_{i,j}\beta p_{j}^t).\nonumber \vspace{-2mm}
	\end{align}
$p_i^t$ is the probability that node $i$ is abnormal at time $t$, $G_{i,j}^{t}$ is the element in the $i$-th row and $j$-th column of $G^t$. The proof is in \ref{apdix:proof}. In each iteration, we calculate $f(k)$ for every node, remove the node with the highest $f(k)$, and update $G^t$, repeating this process $M_2$ times.


\subsection{Model Training and Deployment}

Note that the policy selection module needs to decide whether to call the explore module or the remove module in different scenarios. Thus, when training the policy selection module, the explore module and the remove module should be fixed for the policy selection module to learn the optimal policy. Therefore, during the training process, we train the explore module first and then train the policy selection module. Furthermore, we use the multi-step learning method for the policy selection module in the training stage to make the model converge faster, which can lead to faster learning \cite{sutton1988learning,hessel2018rainbow}.


When deploying our framework, at each time step, the policy selection module determines whether to call the explore module or the remove module. Then the explore/remove module decides which nodes to explore/remove. By adopting this hierarchical reinforcement learning algorithm, the action space for the explore and remove modules is $N$ and that for the policy selection module is 2, which are significantly reduced when compared to $\binom{N}{M_1}+ \binom{N}{M_2}$. 

\section{Experiments}



We run simulations on the synthetic scale-free networks with 100 nodes and an average degree of 4. In the explore module, we employ a three-layer Deep Q network with input, hidden, and output layer dimensions of 5, 80, and 1, respectively. As for the policy selection module, it consists of a three-layer GCN (Graph Convolutional Network) and a four-layer Deep Q network. In the GCN, the input, hidden, and output layer dimensions are 4, 100, and 20, respectively. In the case of the Deep Q network (DQN), each layer has dimensions of 21, 80, 60, and 1, respectively. We compare HIREC with the following baseline methods: \emph{Netshield} \cite{tong2010vulnerability}, \emph{Netshield+} \cite{2015Node2}, \emph{GreedyDrop} \cite{2017Spectral}, \emph{SVID }\cite{2018Group}, and the degree-based method \cite{2015Node}. 
We generate 200 scale-free networks and show the average simulation results below.
Note that existing methods only consider how to control the spread but not how to explore the network.
For a fair comparison, we compare the performance of different algorithms in two different simulation setups. 

In the first simulation setup, note that the baseline methods can only select which nodes to remove, which is equivalent to our proposed remove module. Thus, we fix the policy selection module and the explore module to address the issue of unknown network structure and compare the performance of our proposed remove module with the baseline methods. Specifically, we first train HIREC. Since the three modules in HIREC are independent of each other, we have the flexibility to replace the control module with various baseline methods. These baseline methods can be integrated with the trained policy selection module and exploration module. This allows them to effectively address the spread control problem, even when the network structure is unknown. We set $\beta=0.012$, $\gamma=0.02$ and $T=50$. 
From simulations, we find that if $M_1$ is too small, the amount of information obtained through exploration becomes insufficient, and thus the agent abandons exploration and solely relies on random node removal at each time step. To address this, we set $M_1>M_2$ to ensure more effective exploration of the network structure. We test under different $(M_1,M_2)$ settings, and the results are in Fig. \ref{fig:M1_1} and Fig. \ref{fig:M1_2}. It is evident that our proposed remove module performs significantly better than the baseline methods and reduces the abnormal rate by at least $17\%$.

In the second simulation setup, we assume that the baseline algorithms know a fixed part of the network structure and let $P_r$ denote the percentage of edges they know; while HIREC has no knowledge of the edges in the network at $t=0$. Note that $P_r=100\%$ indicates the baseline algorithms have complete knowledge of the network. Then at each time, these benchmark algorithms select $M_2=1$ node to remove; while HIREC chooses between exploring $M_1$ nodes and removing $M_2=1$ node. 
We set $\beta=0.01$, $\gamma=0.014$ and $T=50$.
The results are in Fig \ref{fig:s2}. First, HIREC can better control the epidemic when we can explore more nodes at one time. Also, the baseline methods perform better when they have more knowledge of the network structure. In addition, when $M_1=15$, HIREC performs better than all baseline methods even when they have complete knowledge of the network structure with $P_r=100\%$. When $M_1=10$ and $M_1=4$, HIREC outperforms the baseline methods when they have information of 80\% and 60\% of the network structure, respectively. These results demonstrate the superior performance of HIREC.

\vspace{-2mm}
\section{Conclusion}

In this paper, we address the challenging problem of controlling the epidemic 
spread over networks with unknown structures, and propose a hierarchical reinforcement learning framework for joint epidemic control and network structure exploration. To reduce the computation complexity, our proposed framework first chooses between the two actions of explore and remove, and then decides which nodes to explore/remove. Simulation results show that our proposed method significantly outperforms existing methods. 


\clearpage

\apptocmd{\thebibliography}{\setlength{\itemsep}{1pt}}{}{}

\bibliographystyle{IEEEbib}
\bibliography{reference.bib}

\clearpage
\appendix
\renewcommand{\thesection}{APPENDIX \Alph{section}}
\section{The proof of Theorem 1}
\label{apdix:proof}
We use $H_i^t$ to represent the state of node $i$ at time $t$. For example, $H_i^t=S$ means that node $i$ is in state $S$ at time $t$. We use $Y_i^{t+1}$ to represent whether the node $i$ becomes abnormal (become exposed) from state $S$ at $t+1$. $Y_i^{t+1}=1$ means the node $i$ become abnormal from state $S$ at $t+1$ and $Y_i^{t+1}=0$ means not. The probability that $Y_i^{t+1}=1$ is:

\begin{align}
P[Y_i^{t+1}=1] &= P \left[H_i^{t+1}=E, H_i^{t} = S \right] \\
&= P \left[H_i^{t+1}=E | H_i^{t} = S \right]P[H_i^{t} = S]\nonumber \\
&= [1-p_{i}^{t}] q_{i}^{t} [1-\prod_{j=1...N}(1-G^{t}_{i,j}\beta p_{j}^{t})].
\end{align}

Since $\delta^{t+1}$ is the number of nodes that become abnormal from state $S$ at time $t+1$, so we have:

\begin{align}
\delta^{t+1} = \sum_{i=1...N}Y_i^{t+1}.
\end{align}

Then the expectation of $\delta^{t+1}$ can be expressed as:

\begin{align}
E(\delta^{t+1}) &= E\left (\sum_{i=1...N}Y_i^{t+1}\right )\nonumber\\
 &=\sum_{i=1...N} E(Y_i^{t+1})\nonumber\\ &=\sum_{i=1...N}\left (1\times P[Y_i^{t+1}=1]+0\times P[Y_i^{t+1}=0]\right )\nonumber \\
 &=\sum_{i=1...N} P[Y_i^{t+1}=1]\nonumber\\
&= \sum_{i=1...N}[1-p_{i}^{t}] q_{i}^{t} [1-\prod_{j=1...N}(1-G^{t}_{i,j}\beta p_{j}^{t})].
\label{eq:exp}
\end{align}

If we remove node $k$ at time $t$, then the expection becomes:

\begin{align}
E'(\delta^{t+1}) &= \sum_{i=1...N,i\neq k}[1-p_{i}^{t}] q_{i}^{t} [1-\prod_{j=1...N,j\neq k}(1-G^{t}_{i,j}\beta p_{j}^{t})].
\end{align}

After remove the node $k$, the reduction of the expection of the $\delta^{t+1}$ is:
\begin{align}
&\Delta E(\delta^{t+1}) = E(\delta^{t+1}) - E'(\delta^{t+1})\nonumber\\
=&(1-p_{k}^t) q_{k}^t [1-\prod_{i=1...N}(1-G^t_{k,i}\beta p_{i}^t)] \nonumber\\
&+\sum_{i=1...N,i\neq k}G^t_{i,k}\beta p_{k}^t(1-p_{i}^t)q_{i}^t\prod_{j=1...N,j\neq k}(1-G^t_{i,j}\beta p_{j}^t).
\end{align}

Since $E'(\delta^{t+1}) = E(\delta^{t+1}) - \Delta E(\delta^{t+1})$, giving $E(\delta^{t+1})$, finding a removed node to minimize $E'(\delta^{t+1})$ is equivalent to find a removed node to maximize $\Delta E(\delta^{t+1})$, that is:

\begin{align}
&\underset{k}{max}\ (1-p_{k}^t) q_{k}^t [1-\prod_{i=1...N}(1-G^t_{k,i}\beta p_{i}^t)] \label{eq:ori}\\
&+\sum_{i=1...N,i\neq k}G^t_{i,k}\beta p_{k}^t(1-p_{i}^t)q_{i}^t\prod_{j=1...N,j\neq k}(1-G^t_{i,j}\beta p_{j}^t).\nonumber
\end{align}

We define
\begin{align}
	&f(k)=(1-p_{k}^t) q_{k}^t [1-\prod_{i=1...N}(1-G^t_{k,i}\beta p_{i}^t)] \label{eq:removeori}\\
	&+\sum_{i=1...N,i\neq k}G^t_{i,k}\beta p_{k}^t(1-p_{i}^t)q_{i}^t\prod_{j=1...N,j\neq k}(1-G^t_{i,j}\beta p_{j}^t).\nonumber \vspace{-2mm}
\end{align}

Then finding the optimal node to remove at $t$ to minimize the expection of $\delta^{t+1}$ is equivalent to finding node $k$ with the highest $f(k)$.

\end{document}